\documentclass[top=0.75in, bottom=1in, conference]{IEEEtran}
\IEEEoverridecommandlockouts
\usepackage{amsmath,amsfonts}
\usepackage{algorithm}
\usepackage{array}
\usepackage{cite}
\usepackage{verbatim}
\usepackage{stfloats}
\usepackage{tabularx}
\usepackage{gensymb}
\usepackage{lipsum}
\usepackage{url}
\usepackage{graphicx}
\usepackage[caption=false,font=normalsize,labelfont=sf,textfont=sf]{subfig}
\usepackage{tikz}
\usetikzlibrary{shapes.geometric, intersections, positioning, backgrounds}

\usepackage{multirow}
\usepackage{colortbl}
\usepackage{hhline}
\usepackage{rotating}
\usepackage[draft,hidelinks,colorlinks=true,citecolor=blue]{hyperref}

\usepackage{colortbl}
\usepackage{pgf}

\usepackage{tabulary}
\usepackage{amssymb,amsthm}
\usepackage{tikz}
\usetikzlibrary{calc}
\usepackage{stackengine}
\usepackage{scalerel}
\usepackage{fp}
\usepackage{color,soul}
\usepackage{enumitem}
\usetikzlibrary{svg.path}
\definecolor{orcidlogocol}{HTML}{A6CE39}
\tikzset{orcidlogo/.pic={
\fill[orcidlogocol] svg{M256,128c0,70.7-57.3,128-128,128C57.3,256,0,198.7,0,128C0,57.3,57.3,0,128,0C198.7,0,256,57.3,256,128z};
\fill[white] svg{M86.3,186.2H70.9V79.1h15.4v48.4V186.2z}
                 svg{M108.9,79.1h41.6c39.6,0,57,28.3,57,53.6c0,27.5-21.5,53.6-56.8,53.6h-41.8V79.1z M124.3,172.4h24.5c34.9,0,42.9-26.5,42.9-39.7c0-21.5-13.7-39.7-43.7-39.7h-23.7V172.4z}
                 svg{M88.7,56.8c0,5.5-4.5,10.1-10.1,10.1c-5.6,0-10.1-4.6-10.1-10.1c0-5.6,4.5-10.1,10.1-10.1C84.2,46.7,88.7,51.3,88.7,56.8z};}}
\newcommand\orcidicon[1]{\href{https://orcid.org/#1}{\mbox{\scalerel*{
\begin{tikzpicture}[yscale=-1,transform shape]
\pic{orcidlogo};
\end{tikzpicture}
}{|}}}}
\theoremstyle{definition}

\usepackage{textcomp}

\usepackage{algpseudocode}
\makeatletter
\newcommand{\heatmap}[1]{%
  \begingroup
  \pgfmathsetmacro{\value}{max(12, min(18.6, #1))} 
  \pgfmathsetmacro{\normval}{(\value - 12)/(18.6 - 12)} 
  \pgfmathsetmacro{\r}{255 * (0.1 + \normval * (1 - 0.1))} 
  \pgfmathsetmacro{\g}{255 * (0.9 - \normval * (0.9 - 0.0))} 
  \pgfmathsetmacro{\b}{255 * (1 - \normval * (1 - 0.5))}     
  \xdef\cellcolorRGB{\r,\g,\b}
  \cellcolor[RGB]{\cellcolorRGB}\phantom{#1}
  \endgroup
}
\makeatother
\makeatletter
\newcommand{\heatmapPower}[1]{%
  \begingroup
  \pgfmathsetmacro{\value}{max(0, min(4200, #1))} 
  \pgfmathsetmacro{\normval}{\value / 4200} 
  \pgfmathsetmacro{\r}{255 * (0.1 + \normval * (1 - 0.1))} 
  \pgfmathsetmacro{\g}{255 * (0.9 - \normval * 0.9)}       
  \pgfmathsetmacro{\b}{255 * (1 - \normval * 0.5)}         
  \xdef\cellcolorRGB{\r,\g,\b}
  \pgfmathsetmacro{\rounded}{int(#1)} 
  \cellcolor[RGB]{\cellcolorRGB}\phantom{\rounded}
  \endgroup
}
\makeatother

\begin{document}
\title{A Lightweight DL Model for Smart Grid Power Forecasting with Feature and Resolution Mismatch}
\author{
\IEEEauthorblockN{
Sarah Al-Shareeda\IEEEauthorrefmark{2}\IEEEauthorrefmark{4}, Gulcihan Ozdemir\IEEEauthorrefmark{3}, Heung Seok Jeon\IEEEauthorrefmark{6}, and Khaleel Ahmad\IEEEauthorrefmark{5}}

\IEEEauthorblockA{\IEEEauthorrefmark{2}BTS Labs, Istanbul Technical University R$\&$D Center, Turkey}

\IEEEauthorblockA{\IEEEauthorrefmark{4}Center for Automotive Research (CAR), The Ohio State University, USA}

\IEEEauthorblockA{\IEEEauthorrefmark{3}Informatics Institute, Istanbul Technical University, Turkey}

\IEEEauthorblockA{\IEEEauthorrefmark{6}Computer Engineering Department, Konkuk University, South Korea}

\IEEEauthorblockA{\IEEEauthorrefmark{5}Computer Science and Information Technology Department, Maulana Azad National Urdu University, India}

{\{alshareeda, ozdemirg\}@itu.edu.tr}, hsjeon@kku.ac.kr, khaleelahmad@manuu.edu.in}
\markboth{}{}
\maketitle

\begin{abstract}
How can short-term energy consumption be accurately forecasted when sensor data is noisy, incomplete, and lacks contextual richness? This question guided our participation in the \textit{2025 Competition on Electric Energy Consumption Forecast Adopting Multi-criteria Performance Metrics}, which challenged teams to predict next-day power demand using real-world high-frequency data. We proposed a robust yet lightweight Deep Learning (DL) pipeline combining hourly downsizing, dual-mode imputation (mean and polynomial regression), and comprehensive normalization, ultimately selecting Standard Scaling for optimal balance. The lightweight GRU-LSTM sequence-to-one model achieves an average RMSE of 601.9~W, MAE of 468.9~W, and 84.36\% accuracy. Despite asymmetric inputs and imputed gaps, it generalized well, captured nonlinear demand patterns, and maintained low inference latency. Notably, spatiotemporal heatmap analysis reveals a strong alignment between temperature trends and predicted consumption, further reinforcing the model's reliability. These results demonstrate that targeted preprocessing paired with compact recurrent architectures can still enable fast, accurate, and deployment-ready energy forecasting in real-world conditions.
\end{abstract}

\IEEEpeerreviewmaketitle

\section{Introduction and Background}
\begin{figure}[!htbp]
\centering
\includegraphics[width=\columnwidth]{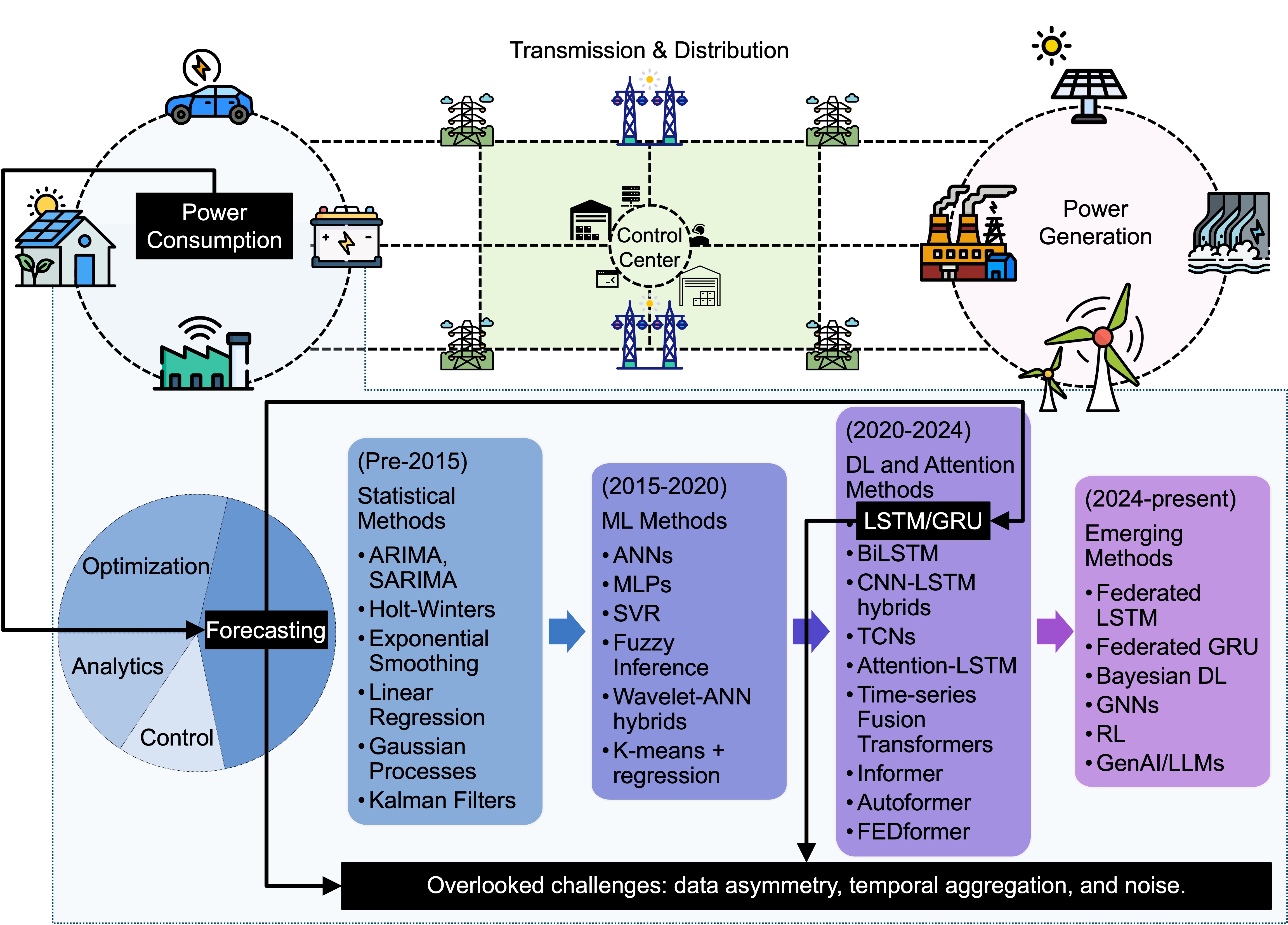}
\caption{Positioning our contribution within the forecasting landscape. Bridging it with real-world data and deployment challenges.}\label{fig:taxonomy}
\end{figure}

How can short-term electric power consumption forecasting models remain accurate, robust, and lightweight when trained on multi-feature, fine-grained data, yet evaluated on single-feature, coarse-grained inputs? This question encapsulates a core challenge in smart grid deployment, where inference-time conditions often deviate significantly from ideal training scenarios. In particular, the problem of asymmetric test-time inputs, characterized by reduced feature availability, lower temporal resolution, and noisy measurements, remains insufficiently addressed in the literature. This paper aims to close that gap. Our work emerges from the ongoing \textit{Competition on Electric Energy Consumption Forecast Adopting Multi-criteria Performance Metrics}, organized by the GECAD Energy and Power Systems Research Group at ISEP, Portugal \cite{gomes_2024_14275645}. The competition is explicitly designed to evaluate model resilience in such asymmetric input settings, emphasizing robustness and deployability under real-world constraints. To contextualize this challenge, we trace the evolution of forecasting approaches across multiple paradigms, as depicted in Fig. \ref{fig:taxonomy}. Classical statistical models such as ARIMA, SARIMA, and exponential smoothing have historically performed well on structured, stationary data, but falter in capturing nonlinear dynamics \cite{ozdemir2024long,ozdemir2024probabilistic}. The subsequent wave of Machine Learning (ML) methods, e.g., Multi-Layer Perceptrons (MLPs), Artificial Neural Networks (ANNs), and Support Vector Regression (SVR), offered greater flexibility, yet remained sensitive to noise and heavily dependent on manual feature engineering \cite{zia2025short, majeed2025data}. Deep Learning (DL) then introduced scalable, end-to-end architectures such as Long Short-Term Memory (LSTM), Gated Recurrent Unit (GRU), and CNN-LSTM hybrids, which substantially improved temporal modeling. More recently, transformer-based models (e.g., Informer, FEDformer), Temporal Convolutional Networks (TCNs), and probabilistic frameworks have advanced long-range prediction capabilities \cite{dakheel2025optimizing}. Parallel efforts in Federated Learning (FL), Bayesian DL, Graph NNs (GNNs), and Reinforcement Learning (RL) aim to address data decentralization, uncertainty quantification, and dynamic adaptability \cite{xu2025review, rahman2025electrical}.

Yet, despite these advancements, the compounded issues of data asymmetry, temporal aggregation, and feature incompleteness at inference time remain underexplored. As underscored in Fig. \ref{fig:taxonomy}, our work specifically targets these overlooked challenges. Motivated by the competition's asymmetric scenario, we propose a forecasting framework that is both lightweight and deployment-ready, tailored to work under degraded input conditions. Our solution combines a hybrid GRU-LSTM architecture with a structured, domain-aware preprocessing pipeline designed to handle mismatched resolutions, missing features, and noisy observations, without resorting to overly complex models or exhaustive tuning. This paper presents our complete Phase I solution, including model design, feature engineering strategies, and experimental evaluation. While final Phase I rankings are still pending, our approach delivers a reproducible, efficient, and practical solution for short-term electric load forecasting in constrained environments. The main contributions of this study are:
\begin{itemize}
\item A unified, lightweight forecasting framework that integrates a hybrid GRU-LSTM architecture with a structured preprocessing pipeline to handle asymmetric inputs, resolution mismatches, and data incompleteness.
\item A comparative evaluation of two imputation strategies (mean-based and polynomial regression) and three normalization schemes (Z-Score, Standard, Min-Max), analyzing their impact on training and predictive accuracy.
\item Empirical validation on the official competition dataset, demonstrating competitive performance, measured by RMSE, MAE, MAPE, and normalized accuracy, even under realistic and degraded test conditions.
\end{itemize}

The work is structured as follows. Section \ref{systemmodel} presents our model design. The simulation settings and results are given in Section \ref{result}. Section \ref{conc} concludes with directions for future research.

\section{Lightweight Power Forecasting Model Design}\label{systemmodel}
\begin{figure*}[!htbp]
    \centering
    \includegraphics[width=.65\linewidth]{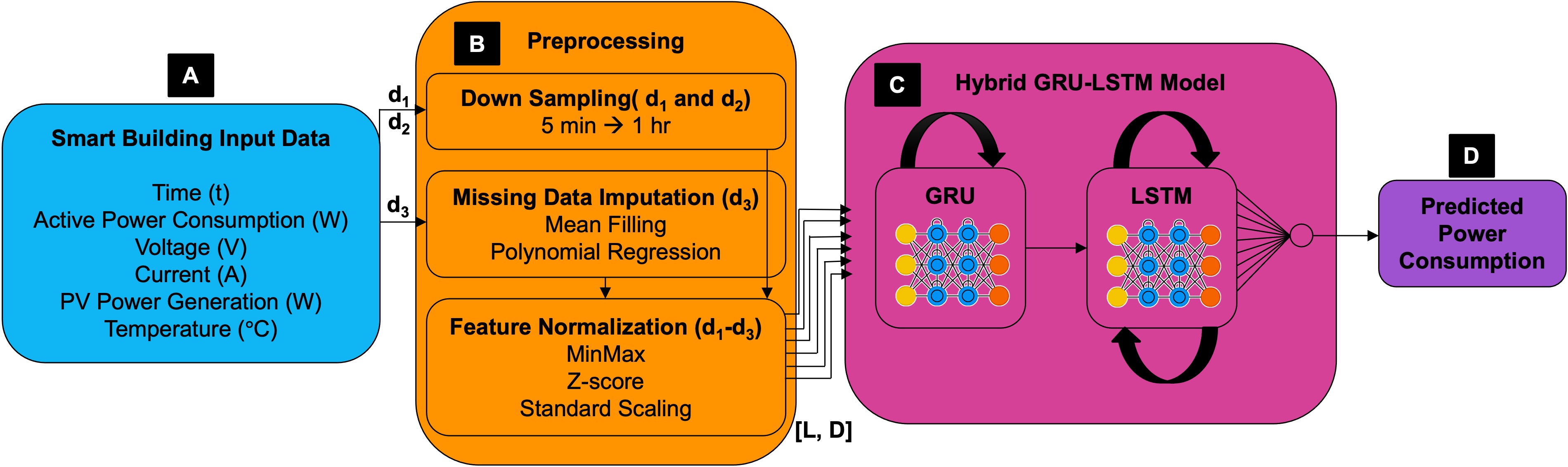}
    \caption{Block diagram of the proposed lightweight forecasting pipeline.}
    \label{fig:blockdiagram}
\end{figure*}

This section proposes our forecasting framework that integrates a structured preprocessing pipeline with a lightweight GRU-LSTM model to address the previously mentioned real-world challenges of feature sparsity, resolution mismatch, and test-time noise. Our system, Fig. \ref{fig:blockdiagram}, is designed to balance accuracy and efficiency while capturing temporal dependencies. Each stage of the pipeline is described below.

\subsection{Input Data Description}
We use the official benchmark dataset from the \textit{2025 Competition on Electric Energy Consumption Forecast Adopting Multi-criteria Performance Metrics} \cite{gomes_2024_14275645}. The dataset comprises high-resolution time-series measurements from a smart building and is divided into three temporally disjoint subsets:
\begin{itemize}
\item $d_1$: One year of 5-minute interval data (01/12/2023–01/12/2024) used exclusively for training, capturing long-term trends and seasonality.
\item $d_2$: A 40-day sequence (starting 29/11/2024) at 5-minute resolution, used for validation and hyperparameter tuning.
\item $d_3$: The competition test set covering five days (06–10/01/2025), where hourly temperature forecasts are released in advance, followed by the corresponding ground-truth consumption data.
\end{itemize}

Subsets $d_1$ and $d_2$ contain the following six variables: Timestamp $t$ (min, hr), Power consumption $P$ (W), Voltage $V$ (V), Current $I$ (A), PV generation $P_{PV}$ (W), and Temperature $T$ (\degree C). Given its real-world origin, the dataset includes missing values, noise, and occasional anomalies. A robust preprocessing pipeline is therefore required to ensure effective learning, as detailed in the next subsection.

\subsection{Data Preprocessing}
To handle the high-frequency nature and imperfections of the dataset, we design a three-stage preprocessing pipeline: (i) downsampling of training and validation data ($d_1$, $d_2$) via mean aggregation to match the hourly resolution of $d_3$; (ii) imputation of missing values in the test data using both mean-based and third-order polynomial regression techniques; (iii) normalization using Z-Score, Standard, and Min-Max scaling to enhance training stability and ensure feature comparability.

\subsubsection{Downsizing of Training and Validation Data}
The training ($d_1$) and validation ($d_2$) datasets are recorded at 5-minute intervals, while the test set ($d_3$) is at hourly resolution. To ensure temporal alignment and reduce noise, we downsample $d_1$ and $d_2$ using mean aggregation over every 12 consecutive samples. For any univariate signal $x_t$, the hourly value is computed as $x'_k = \frac{1}{12} \sum_{i=0}^{11} x_{12k + i}$. This transformation is applied independently to all features in $d_1$ and $d_2$.

\subsubsection{Imputation of Test Data}
The test dataset $d_3$ is intentionally incomplete, providing only temperature $T$ and timestamp $t$. To reconstruct the missing features\textcolor{black}{($V$, $I$, $P_{PV}$)}, we employ two complementary imputation strategies:
\begin{itemize}
    \item \textcolor{black}{Mean-based filling: Each missing feature in ($V$, $I$, $P_{PV}$) is replaced by the average of its peer value computed from the training set $d_1$.}
    \item Polynomial regression: A 3rd-degree polynomial model is fit to each missing variable using temperature $T$ as the predictor. The model is trained on $d_1$ data, and the imputed value $\hat{y}$ is given by $\hat{y} = \beta_0 + \beta_1 T + \beta_2 T^2 + \beta_3 T^3$, where $\beta_0, \beta_1, \beta_2, \beta_3$ are learned coefficients. This method captures non-linear relationships between temperature and the target feature, providing a more context-aware reconstruction than simple mean filling.
\end{itemize}

\subsubsection{Normalization Step}
To ensure training stability and balanced feature contributions, we normalize both the input features ${X}$ and the target variable $P$ across all datasets using three techniques:
\begin{itemize}
    \item Standard scaling: each feature \textit{x} is transformed to zero mean and unit variance using mean $\mu$ and standard deviation $\sigma$ from $d_1$.
    \item Min-max scaling: each feature is rescaled to the $[0, 1]$ range using the minimum and maximum values from $d_1$.
    \item manual Z-Score normalization: a custom transformation $x^* = \frac{x - \mu}{\sigma}$ is applied to offer reproducibility.
\end{itemize}
All transformations are consistently applied to ${X}$ and $P$ using statistics from $d_1$. For accurate prediction recovery, the normalization parameters for $P$ (i.e., $\mu_P$, $\sigma_P$) are retained and reused during inverse transformation. With the data now downsampled, imputed, and normalized, we proceed to train our GRU-LSTM model for short-term power forecasting.

\subsection{Hybrid GRU-LSTM Forecasting Model}
The goal of the proposed model is to forecast short-term power consumption using recent sensor observations. At each time step $t$, the model takes as input a sequence of $L$ past observations of the feature vector ${x}_t \in \mathbb{R}^D$ and outputs a one-step-ahead prediction $\hat{P}_{t+1}$ of the target variable $P$. The forecasting function $f$ is defined as:
\begin{equation}
\hat{P}_{t+1} = f( {x}_{t-L+1}, \dots, {x}_t )
\end{equation}
where $L$ denotes the sequence length, and $D$ is the dimensionality of the input vector.

For the training and validation datasets ($d_1$ and $d_2$), the feature vector ${x}_t$ includes the full set of six variables:
\begin{equation}
{x}_t = [ t,\; P,\; V,\; I,\; P_{PV},\; T ]^\top
\end{equation}
where $t$ is the timestamp, $P$ is the power consumption target, $V$ is voltage, $I$ is current, $P_{PV}$ is PV generation, and $T$ is temperature. For the feature-asymmetric test dataset $d_3$, only $t$ and $T$ are available. Therefore, the missing features in ${x}_t$ are reconstructed using imputation strategies to bridge the train-test feature gap, detailed in the preprocessing section. After completion and normalization, the model receives input vectors of the same dimension $D$ across all datasets.

\subsubsection{GRU-LSTM Architecture}
Our forecasting model adopts a hybrid recurrent architecture combining a Bidirectional GRU (BiGRU) with a unidirectional LSTM to capture both short-term fluctuations and long-range dependencies in our asymmetric data. The model is structured as follows:

\begin{itemize}
    \item BiGRU layer: A Bidirectional GRU with 256 hidden units and ReLU activation processes input sequences in both forward and backward directions. This enhances context inference, particularly when the input window length \( L \) is small. The GRU's simplified gating (update and reset gates) ensures computational efficiency while capturing variations in power and environmental variables.
    \item dropout layer: A dropout layer with rate \( r \) follows to prevent overfitting by randomly deactivating neurons during training, encouraging more robust feature representations.
    \item LSTM layer: The BiGRU output feeds into a unidirectional LSTM with 128 hidden units and ReLU activation. The LSTM's internal memory cell captures longer-range dependencies, such as delayed effects of temperature on energy demand, through its gates.
    \item second dropout layer: An additional dropout layer with the same rate \( r \) enhances generalization by reducing neuron co-adaptation prior to final prediction.
    \item output layer: A fully connected single-neuron layer maps the LSTM's final hidden state to the one-step-ahead forecast \( \hat{P} \). This scalar output is inverse-transformed using preserved normalization parameters to recover the original scale of the target variable \( P \).
\end{itemize}

This hybrid architecture combines GRU's efficiency and short-term sensitivity with LSTM's long-term memory capacity, supporting accurate forecasting in dynamic, sensor-driven environments with incomplete or imputed features.

\subsubsection{Model Training and Validation}
The hybrid GRU-LSTM model is trained separately for each normalized variant, Standard scaling, Min-Max scaling, and Z-score normalization, using the corresponding preprocessed training set $d_1$ and validation set $d_2$. For each variant, a sliding window of length $L$ is applied to generate supervised sequence-to-one prediction samples. At each time step $t$, the model receives a multivariate input sequence $\{{x}_{t-L+1}, \dots, {x}_t\}$, where ${x}_t \in \mathbb{R}^D$ includes all $D$ features: $t$, $P$, $V$, $I$, $P_{PV}$, and $T$, reflecting the complete feature availability during training. The learning objective is to predict the power consumption $P$ at the next time step, $y_{t+1} = P_{t+1}$, using the function $\hat{P}_{t+1} = f( {x}_{t-L+1}, \dots, {x}_t)$. Each training sample can thus be expressed as $( {X}_t, y_{t+1} ) \in ( \mathbb{R}^{L \times D},\; \mathbb{R})$.

Training is performed over a maximum of 200 epochs with a batch size of 16. To prevent overfitting, an early stopping mechanism monitors validation loss on the external validation set $d_2$ and halts training if no improvement is observed over 5 consecutive epochs, restoring the best-performing weights. A learning rate scheduler further enhances convergence by reducing the learning rate when validation loss stagnates for 3 epochs, with a lower bound set to $10^{-6}$. This external validation setup, distinct from internal splits, ensures an unbiased estimate of generalization performance and supports consistent hyperparameter tuning across normalization variants. Following training, the model is assessed on the test set $d_3$, with performance results and analysis presented in the next section.

\section{Simulations and Analysis}\label{result}
This section presents the experimental configuration, dataset usage, model training, and forecasting results under different normalization schemes. All experiments are conducted using Python 3.11.3 in JupyterLab on a MacBook Pro equipped with a 2.8 GHz Intel Core i7 CPU and 16 GB RAM \cite{10910975,citation-0,10323976}. The trained GRU-LSTM model is evaluated on five distinct days (06/01/2025 - 10/01/2025) from the $d_3$ test set, with each day comprising 24 hourly time steps, i.e., $N = 24$. In accordance with the competition protocol \cite{gomes_2024_14275645}, each test file initially provides only the temperature $T$ at timestamp $t$ for the forecast horizon ${t+1, \dots, t+N}$. To enable forecasting under this asymmetric input setting, the full feature vector ${x}_t \in \mathbb{R}$ is first reconstructed via imputation using statistical relationships learned from the training set $d_1$. These imputed features are then normalized using the same scaling parameters $(\mu, \sigma)$ derived from $d_1$, ensuring consistency across datasets. The model processes the reconstructed input sequence ${{x}_{t-L+1}, \dots, {x}_t}$ and produces a sequence of one-step-ahead predictions ${\hat{P}{t+1}, \dots, \hat{P}{t+N}}$. These predicted values are subsequently inverse-transformed using $(\mu_P, \sigma_P)$ to restore the original scale of the target variable $P$. After the initial forecasting, the true values ${P{t+1}, \dots, P_{t+N}}$ are released by the organizers, enabling final evaluation.

\subsection{Evaluation Metrics}
To quantitatively assess the forecasting accuracy of the GRU-LSTM model, we compare the predicted power values $\{\hat{P}_{t+1}, \dots, \hat{P}_{t+N}\}$ with the ground truth values $\{P_{t+1}, \dots, P_{t+N}\}$ over each test day, where $N = 24$ is the number of hourly time steps in the prediction horizon. The following evaluation metrics are considered:

\begin{itemize}
    \item RMSA that measures the standard deviation of the prediction errors and penalizes larger deviations more heavily:
    \begin{equation}
        \text{RMSE} = \sqrt{ \frac{1}{N} \sum_{i=1}^{N} \left(P_{t+i} - \hat{P}_{t+i}\right)^2 }
    \end{equation}

    \item MAE that captures the average absolute magnitude of forecasting errors, independent of direction:
    \begin{equation}
        \text{MAE} = \frac{1}{N} \sum_{i=1}^{N} \left| P_{t+i} - \hat{P}_{t+i} \right|
    \end{equation}
    \item MAPE to evaluate the prediction error relative to the actual value, expressed as a percentage. A small constant $\epsilon$ is used to avoid division by zero:
    \begin{equation}
        \text{MAPE} = \frac{100}{N} \sum_{i=1}^{N} \left| \frac{P_{t+i} - \hat{P}_{t+i}}{\max(|P_{t+i}|, \epsilon)} \right|, \quad \epsilon > 0
    \end{equation}
    \item Normalized accuracy (\%) provides a percentage-based estimate of prediction precision relative to the dynamic range of the target variable:
    \begin{equation}
    \text{Accuracy} = 100 \times ( 1 - \frac{ \frac{1}{N} \sum_{t=1}^{N} | P_t - \hat{P}_t | }{ \max(P_t) - \min(P_t) } )
\end{equation}
    \item Prediction computation latency measures the inference time (sec) required to generate the 24-step forecast for each test instance.
\end{itemize}

To qualitatively assess the model, its ability to track temporal trends and accurately capture consumption magnitudes, below we further illustrate all forecasting results by comparing predicted values $\hat{P}{t+i}$ against actual observations $P{t+i}$ across the full forecast horizon.

\subsection{Results and Discussions}
This subsection reports forecasting results for the trained GRU–LSTM on five test days in \(d_3\). For each day, hourly consumption \(\hat{P}_t\) is predicted from imputed, normalized inputs and inverse-transformed for comparison with ground truth \(P_t\) released by the competition organizers \cite{gomes_2024_14275645}. Performance is evaluated using RMSE, MAE, MAPE, accuracy, and prediction latency under three normalization strategies (Table~\ref{tab:results_compact}). Standard scaling yields the best overall results, with the lowest RMSE (e.g., 518.8~W on Day~3), lowest MAPE (13.2\% on Day~2), and highest accuracy (up to 86.8\%). Z-Score normalization is comparable and surpasses standard scaling on Days~1 and 5 in RMSE (e.g., 572.6~W vs. 721.1~W on Day~5). Min-Max scaling performs notably worse, with RMSE often above 1500~W, MAPE above 40\%, and accuracy as low as 56.8\%, highlighting the limits of compressing features to \([0,1]\) under distribution shift.

Prediction latencies remain low across all strategies (0.065 to 0.17~sec), supporting real-time deployment. The model maintains stable performance across all test days without retraining, indicating strong generalization. Under Standard scaling the model tracks daily demand patterns, including morning peaks between 06:00 AM and 12:00 PM and subsequent stable periods; minor deviations align with abrupt changes or missing contextual features. The close agreement between \(\hat{P}_t\) and \(P_t\) under Standard scaling supports the model's reliability for short-term load forecasting in data-limited settings. To examine the spatiotemporal correlation between real temperature ($T$) and predicted consumption ($P$), the heatmaps in Table~\ref{tab:hmt} and Table~\ref{tab:hmp} compare five consecutive days (06/01–10/01/2025). The $T$ heatmap exhibits consistent diurnal patterns, with early morning lows (e.g., 11.7$^\circ$C on 07/01) and afternoon peaks (up to 18.6$^\circ$C on 10/01), reflecting typical thermal behavior. In contrast, the $P$ heatmap reveals temporal variability and amplitude, with daytime peaks, highlighting the model's sensitivity to occupancy and appliance use. Day 5 shows a slight mismatch between $T$ and $P$, aligning with the relatively lower accuracy reported in Table \ref{tab:results_compact} and low-activity hence low consumption in such smart building. Despite asymmetric and imputed inputs, the lightweight GRU-LSTM model reliably captures temporal dependencies and delivers robust short-term forecasts, demonstrating strong generalization and readiness for real-time deployment.

\begin{table}[!htbp]
\caption{Forecasting Performance on $d_3$ Test Data Across Five Days}
\label{tab:results_compact}
\centering
\resizebox{.85\columnwidth}{!}{
\begin{tabular}{|c|c|m{.8cm}|m{.8cm}|m{.8cm}|m{.8cm}|m{.8cm}|}\hline
\textbf{Day} & \textbf{Norm} & \textbf{RMSE (W) $\downarrow$} & \textbf{MAE (W) $\downarrow$} & \textbf{MAPE (\%) $\downarrow$} & \textbf{Acc (\%) $\uparrow$} & \textbf{Pred. Latency (sec) $\downarrow$} \\\hline
\multirow{3}{*}{1} 
& Standard   & 630.8 & 481.8 & 16.1 & 83.9 & 0.07 \\
& Min-Max    & 1514.4 & 1227.9 & 40.9 & 59.1 & 0.13 \\
& Z-Score    & 653.6 & 492.9 & 16.4 & 83.6 & 0.11 \\\hline
\multirow{3}{*}{2} 
& Standard   & 523.3 & 397.5 & 13.2 & 86.8 & 0.11 \\
& Min-Max    & 1216.2 & 1030.7 & 34.4 & 65.6 & 0.10 \\
& Z-Score    & 504.9 & 394.8 & 13.2 & 86.8 & 0.17 \\\hline
\multirow{3}{*}{3} 
& Standard   & 518.8 & 409.5 & 13.7 & 86.3 & 0.12 \\
& Min-Max    & 1454.0 & 1260.5 & 42.0 & 58.0 & 0.12 \\
& Z-Score    & 488.2 & 394.0 & 13.1 & 86.9 & 0.07 \\\hline
\multirow{3}{*}{4} 
& Standard   & 611.5 & 467.9 & 15.6 & 84.4 & 0.16 \\
& Min-Max    & 1539.3 & 1266.6 & 42.2 & 57.8 & 0.10 \\
& Z-Score    & 607.4 & 471.4 & 15.7 & 84.3 & 0.14 \\\hline
\multirow{3}{*}{5} 
& Standard   & 721.1 & 587.0 & 19.6 & 80.4 & 0.11 \\
& Min-Max    & 1604.2 & 1295.9 & 43.2 & 56.8 & 0.13 \\
& Z-Score    & 572.6 & 484.7 & 16.2 & 83.8 & 0.07 \\\hline
\end{tabular}
}
\end{table}

\begin{table}[!htbp]
\centering
\caption{Real Temperature Values from 06/01/2025 to 10/01/2025. \textcolor{cyan}{Blue} represents low $T$s and \textcolor{magenta}{Pink-Red} represents high $T$s.}
\resizebox{.72\columnwidth}{!}{
\begin{tabular}{|c|c|c|c|c|c|}
\hline
\textbf{Time} & \textbf{06/01} & \textbf{07/01} & \textbf{08/01} & \textbf{09/01} & \textbf{10/01} \\
\hline
1:00 AM & \heatmap{13.2} & \heatmap{11.7} & \heatmap{13.3} & \heatmap{15.7} & \heatmap{16.1} \\
2:00 AM & \heatmap{13.2} & \heatmap{12.6} & \heatmap{13.7} & \heatmap{15} & \heatmap{16.1} \\
3:00 AM & \heatmap{12.9} & \heatmap{12.7} & \heatmap{13.1} & \heatmap{14.4} & \heatmap{16.5} \\
4:00 AM & \heatmap{12.2} & \heatmap{12.5} & \heatmap{14.2} & \heatmap{14.4} & \heatmap{16.4} \\
5:00 AM & \heatmap{12} & \heatmap{12.7} & \heatmap{14.1} & \heatmap{14.2} & \heatmap{16.4} \\
6:00 AM & \heatmap{12} & \heatmap{13.2} & \heatmap{13.7} & \heatmap{13.7} & \heatmap{15.7} \\
7:00 AM & \heatmap{12.2} & \heatmap{13.1} & \heatmap{14.6} & \heatmap{13.3} & \heatmap{15.5} \\
8:00 AM & \heatmap{12.2} & \heatmap{13.4} & \heatmap{13.9} & \heatmap{13.1} & \heatmap{15.1} \\
9:00 AM & \heatmap{13} & \heatmap{14} & \heatmap{13.8} & \heatmap{13.4} & \heatmap{15.2} \\
10:00 AM & \heatmap{14} & \heatmap{14.8} & \heatmap{14.4} & \heatmap{14.5} & \heatmap{15.4} \\
11:00 AM & \heatmap{14.3} & \heatmap{15.4} & \heatmap{14.2} & \heatmap{15} & \heatmap{16.5} \\
12:00 PM & \heatmap{14.2} & \heatmap{16.1} & \heatmap{15} & \heatmap{15.6} & \heatmap{16.9} \\
1:00 PM & \heatmap{15.7} & \heatmap{17.1} & \heatmap{14.9} & \heatmap{17.4} & \heatmap{17.5} \\
2:00 PM & \heatmap{16.4} & \heatmap{17.2} & \heatmap{15.7} & \heatmap{16.7} & \heatmap{17.1} \\
3:00 PM & \heatmap{15.8} & \heatmap{16.8} & \heatmap{16.1} & \heatmap{17.5} & \heatmap{16.6} \\
4:00 PM & \heatmap{16.1} & \heatmap{16.9} & \heatmap{15.9} & \heatmap{16.7} & \heatmap{16.2} \\
5:00 PM & \heatmap{15.7} & \heatmap{16.8} & \heatmap{16.6} & \heatmap{17} & \heatmap{16.8} \\
6:00 PM & \heatmap{15.1} & \heatmap{16} & \heatmap{16.7} & \heatmap{16.1} & \heatmap{17.4} \\
7:00 PM & \heatmap{14.4} & \heatmap{15.9} & \heatmap{16.9} & \heatmap{16.5} & \heatmap{17.1} \\
8:00 PM & \heatmap{13.4} & \heatmap{15.6} & \heatmap{17.2} & \heatmap{16.6} & \heatmap{18.2} \\
9:00 PM & \heatmap{13.3} & \heatmap{15.1} & \heatmap{17.1} & \heatmap{16.4} & \heatmap{18.6} \\
10:00 PM & \heatmap{13} & \heatmap{14} & \heatmap{17.4} & \heatmap{16.3} & \heatmap{17.4} \\
11:00 PM & \heatmap{13} & \heatmap{13.9} & \heatmap{16.7} & \heatmap{15.7} & \heatmap{18.3} \\
12:00 AM & \heatmap{12.5} & \heatmap{13.4} & \heatmap{16.3} & \heatmap{16.2} & \heatmap{18.1} \\\hline
\end{tabular}}\label{tab:hmt}
\end{table}

\begin{table}[!htbp]
\centering
\caption{Predicted Power Consumption (W) from 06/01/2025 to 10/01/2025. \textcolor{cyan}{Blue} represents low consumption $P$s and \textcolor{magenta}{Pink-Red} represents high consumption $P$s.}
\label{tab:predicted_power_heatmap}
\resizebox{.73\columnwidth}{!}{
\begin{tabular}{|c|c|c|c|c|c|}
\hline
\textbf{Time} & \textbf{06/01} & \textbf{07/01} & \textbf{08/01} & \textbf{09/01} & \textbf{10/01} \\
\hline
1:00 AM & \heatmapPower{1537.647} & \heatmapPower{-290.343} & \heatmapPower{366.007} & \heatmapPower{1675.498} & \heatmapPower{917.337} \\
2:00 AM & \heatmapPower{1537.647} & \heatmapPower{694.838} & \heatmapPower{700.728} & \heatmapPower{1232.714} & \heatmapPower{917.337} \\
3:00 AM & \heatmapPower{1507.615} & \heatmapPower{786.295} & \heatmapPower{134.446} & \heatmapPower{956.764} & \heatmapPower{1615.044} \\
4:00 AM & \heatmapPower{1012.036} & \heatmapPower{597.643} & \heatmapPower{954.534} & \heatmapPower{956.764} & \heatmapPower{1432.643} \\
5:00 AM & \heatmapPower{829.361} & \heatmapPower{786.295} & \heatmapPower{911.116} & \heatmapPower{852.657} & \heatmapPower{1432.643} \\
6:00 AM & \heatmapPower{829.361} & \heatmapPower{1005.433} & \heatmapPower{700.728} & \heatmapPower{513.597} & \heatmapPower{339.385} \\
7:00 AM & \heatmapPower{1012.036} & \heatmapPower{982.366} & \heatmapPower{1103.803} & \heatmapPower{83.213} & \heatmapPower{90.982} \\
8:00 AM & \heatmapPower{1012.036} & \heatmapPower{1058.050} & \heatmapPower{814.413} & \heatmapPower{-198.777} & \heatmapPower{-341.118} \\
9:00 AM & \heatmapPower{1526.741} & \heatmapPower{1200.225} & \heatmapPower{760.428} & \heatmapPower{209.199} & \heatmapPower{-239.958} \\
10:00 AM & \heatmapPower{1468.943} & \heatmapPower{1368.598} & \heatmapPower{1030.608} & \heatmapPower{1005.394} & \heatmapPower{-24.478} \\
11:00 AM & \heatmapPower{1512.117} & \heatmapPower{1655.418} & \heatmapPower{954.534} & \heatmapPower{1232.714} & \heatmapPower{1615.044} \\
12:00 PM & \heatmapPower{1491.602} & \heatmapPower{2659.265} & \heatmapPower{1270.912} & \heatmapPower{1587.573} & \heatmapPower{2604.150} \\
1:00 PM & \heatmapPower{2828.752} & \heatmapPower{3421.583} & \heatmapPower{1223.890} & \heatmapPower{3339.157} & \heatmapPower{3119.832} \\
2:00 PM & \heatmapPower{3659.774} & \heatmapPower{3459.990} & \heatmapPower{1906.927} & \heatmapPower{3056.275} & \heatmapPower{2836.004} \\
3:00 PM & \heatmapPower{2961.481} & \heatmapPower{3308.158} & \heatmapPower{2868.460} & \heatmapPower{3378.065} & \heatmapPower{1839.949} \\
4:00 PM & \heatmapPower{3341.465} & \heatmapPower{3345.479} & \heatmapPower{2429.273} & \heatmapPower{3056.275} & \heatmapPower{1082.021} \\
5:00 PM & \heatmapPower{2828.752} & \heatmapPower{3308.158} & \heatmapPower{3124.164} & \heatmapPower{3175.012} & \heatmapPower{2412.358} \\
6:00 PM & \heatmapPower{2029.562} & \heatmapPower{2483.535} & \heatmapPower{3154.879} & \heatmapPower{2647.949} & \heatmapPower{3045.586} \\
7:00 PM & \heatmapPower{1540.387} & \heatmapPower{2314.322} & \heatmapPower{3221.301} & \heatmapPower{3000.327} & \heatmapPower{2836.004} \\
8:00 PM & \heatmapPower{1521.985} & \heatmapPower{1812.382} & \heatmapPower{3330.665} & \heatmapPower{3024.928} & \heatmapPower{3740.365} \\
9:00 PM & \heatmapPower{1534.890} & \heatmapPower{1486.246} & \heatmapPower{3293.745} & \heatmapPower{2982.022} & \heatmapPower{4138.178} \\
10:00 PM & \heatmapPower{1526.741} & \heatmapPower{1200.225} & \heatmapPower{3403.040} & \heatmapPower{2963.930} & \heatmapPower{3045.586} \\
11:00 PM & \heatmapPower{1526.741} & \heatmapPower{1183.064} & \heatmapPower{3154.879} & \heatmapPower{1675.498} & \heatmapPower{3836.738} \\
12:00 AM & \heatmapPower{1526.741} & \heatmapPower{1183.064} & \heatmapPower{3154.879} & \heatmapPower{1675.498} & \heatmapPower{3836.738} \\
\hline
\end{tabular}}\label{tab:hmp}
\end{table}

\section{Conclusion and Future Work}\label{conc}
This paper introduced a robust forecasting pipeline for power consumption that effectively handles missing data, asymmetric sampling, and limited contextual inputs. Trained on one year of high-frequency data and validated over a 40-day period, the model was deployed to predict next-day consumption across five target days from the \textit{2025 Competition on Electric Energy Consumption Forecast Adopting Multi-criteria Performance Metrics} dataset \cite{gomes_2024_14275645}. The pipeline integrates hourly downsizing, dual-mode imputation, and multiple normalization strategies with a lightweight GRU-LSTM architecture. Using standard scaling, the model achieved an average RMSE of 601.9W, MAE of 468.9W, and 84.36\% accuracy. Despite imputed and asymmetric inputs, the model generalized well, maintained low inference latency, and captured nuanced load dynamics, evident in its ability to distinguish similar temperature profiles (e.g., 06/01/2025 vs. 08/01/2025). A minor accuracy drop on Day 5, linked to input inconsistencies and expected nighttime inactivity, underscored the model's sensitivity to data quality. These results affirm that targeted preprocessing and compact DL architectures can deliver accurate, fast, and generalizable forecasts for real-time energy management. Future work will extend this framework to multi-step, uncertainty modeling, and attention-based enhancements.

\section*{Acknowledgment}
This work was supported by the Research Fund of the Istanbul Technical University. Project Number: 47198.
\bibliographystyle{unsrt}
\bibliography{main}
\end{document}